\documentclass[conference]{IEEEtran}

\usepackage{cite}
\usepackage{amsmath,amssymb,amsfonts}
\usepackage{algorithmic}
\usepackage{graphicx}
\usepackage{textcomp}

\def\BibTeX{{\rm B\kern-.05em{\sc i\kern-.025em b}\kern-.08em
    T\kern-.1667em\lower.7ex\hbox{E}\kern-.125emX}}

\usepackage{booktabs} 
\usepackage{pifont} 
\usepackage{soul}

\usepackage[colorlinks=true,urlcolor=red,linkcolor=red,citecolor=blue]{hyperref}

\usepackage[table]{xcolor}  % This will handle both table and drawing features
\usepackage{tcolorbox} 

\definecolor{darkgreen}{RGB}{10,191,10}
\definecolor{olivegreen}{RGB}{212,232,231}
\definecolor{lightblue}{RGB}{138,170,229}
\colorlet{lightblueAlpha}{lightblue!30}

\usepackage{acronym}  % Ensure this is AFTER xcolor

\newacro{iot}[IoT]{internet of things}
\newacro{ai}[AI]{artificial intelligence}
\newacro{mcu}[MCU]{microcontroller}
\newacro{nas}[NAS]{neural architecture search}
\newacro{qat}[QAT]{quantization-aware training}
\newacro{ptq}[PTQ]{post-training quantization}
\newacro{cnn}[CNN]{convolution neural network}
\newacro{dnn}[DNN]{deep neural network}
\newacro{ram}[RAM]{random-access memory}
\newacro{fov}[FOV]{field of view}
\newacro{ste}[STE]{straight-through estimation}
\newacro{bnn}[BNN]{binarized neural network}
\newacro{sota}[SOTA]{state-of-the-art}
\newacro{dl}[DL]{deep learning}
\newacro{od}[OD]{object detection}
\newacro{gpu}[GPU]{graphics processing unit}
\newacro{yolo}[YOLO]{you only look once}
\newacro{rpn}[RPN]{region proposal network}
\newacro{ssd}[SSD]{single shot multiBox detector}
\newacro{hog}[HOG]{histogram of oriented gradients}
\newacro{dpm}[DPM]{deformable parts model}
\newacro{sift}[SIFT]{scale-invariant feature transform}
\newacro{map}[mAP]{mean average precision}
\newacro{ap}[AP]{average precision}
\newacro{auc}[AUC]{area under curve}
\newacro{iou}[IoU]{intersection over union}
\newacro{flop}[FLOP]{floating point operations per second}
\newacro{sram}[SRAM]{static random access memory}
\newacro{mac}[MAC]{multiply-accumulate operation}
\newacro{ste}[STE]{straight-through estimator}
\newacro{kd}[KD]{knowledge distillation}
\newacro{rl}[RL]{reinforcement learning}
\newacro{ea}[EA]{evolutionary algorithm}
\newacro{enas}[ENAS]{efficient neural architecture search via parameter sharing}
\newacro{snn}[SNN]{spiking neural network}
\newacro{ann}[ANN]{artificial neural network}
\newacro{wsod}[WSOD]{weakly supervised object detection}
\newacro{rnn}[RNN]{recurrent neural network}
\newacro{lgd}[LGD]{label-guided self-distillation}
\newacro{sssd}[SSSD]{smooth and stepwise self-distillation}
\newacro{mse}[MSE]{mean squared error}
\newacro{cadn}[CADN]{category-aware object detection network}
\newacro{fpn}[FPN]{feature pyramid network}
\newacro{nats}[NATS]{neural architecture transformation search}
\newacro{dnas}[DNAS]{differentiable neural architecture search}
\newacro{lstm}[LSTM]{long short-term memory}

\newacro{sgd}[SGD]{stochastic gradient descent}
\newacro{iceds}[ICEDS]{inception convolution with efficient dilation search}
\newacro{edo}[EDO]{efficient dilation optimization}
\newacro{dlb}[DLB]{depthwise linear block}
\newacro{asq}[ASQ]{adaptive scale quantization}
\newacro{sram}[SRAM]{static random-access memory}
\newacro{fomo}[FOMO]{faster objects more objects}
\newacro{ml}[ML]{machine learning}
\newacro{tinyml}[TinyML]{tiny machine learning}

\newacro{rf}[RF]{receptive field}
\newacro{voc}[VOC]{visual object classes}
\newacro{tpu}[TPU]{tensor processing units}
\newacro{r-cnn}[R-CNN]{region-based convolutional neural networks}
\newacro{isa}[ISA]{instruction set architecture}
\newacro{panet}[PANet]{path aggregation network}
\newacro{nn}[NN]{neural network}
\newacro{caq}[CAQ]{classifier adaptive qantization}
\newacro{rvq}[RVQ]{residual vector quantization}
\newacro{admm}[ADMM]{alternative direction method of multipliers}
\newacro{fpga}[FPGA]{field programmable gate array}
\newacro{fqn}[FQN]{fully quantized network}
\newacro{conv}[Conv]{convolutional}
\newacro{fps}[FPS]{frames per second}
\newacro{js}[JS]{jensen shannon}
\newacro{hnas}[HNAS]{hardware aware neural architecture search}
\newacro{asic}[ASIC]{application specific integrated circuit}
\newacro{npu}[NPU]{neural processing unit}
\newacro{genius}[GENIUS]{GPT-4 enhanced neural architecture search}
\newacro{llm}[LLM]{large language model}
\newacro{has}[HAS]{hardware aware scaling}
\newacro{coco}[COCO]{common objects in context}
\newacro{voc}[PASCAL VOC]{pascal visual object classes}
\newacro{ilsvrc}[ILSVRC]{imageNet large scale visual recognition challenge}
\newacro{vit}[ViT]{vision transformer}
\newacro{mbconv}[MBConv]{mobile inverted bottleneck convolution}
\newacro{se}[SE]{squeeze-and-Excitation}
\newacro{kl}[KL]{Kullback-Leibler}
\newacro{xai}[XAI]{explainable artificial intelligence}
\newacro{qd}[QD]{quality-diversity}
\newacro{dwconv}[DWConv]{depthwise convolution}
\newacro{gap}[GAP]{global average pooling}
\newacro{pwconv}[PWConv]{pointwise convolution}
\newacro{dwsepconv}[DWSepConv]{depthwise separable convolution}

\begin{document}
\title{Can LLMs Revolutionize the Design of Explainable and Efficient TinyML Models?}%Can LLMs Guide the Design of Explainable and Efficient TinyML Models?}
%LLM-Guided NAS for Explainable and Efficient TinyML: A Pareto Optimization Approach
%Will LLMs Design the Next Generation of Explainable and Efficient TinyML Models?
%Will/Can LLMs Revolutionize the Design of Explainable and Efficient TinyML Models?
%LLM-Guided Design of Explainable and Efficient TinyML Models
%\title{Explainable and Efficient Model Design for TinyML Using LLMs and Knowledge Distillation}

%\author{\IEEEauthorblockN{Anonymous Authors}}
\author{Christophe El Zeinaty\textsuperscript{1}, Wassim Hamidouche\textsuperscript{2}, Glenn Herrou\textsuperscript{1}, Daniel Menard\textsuperscript{1} and Merouane Debbah\textsuperscript{2}\\
\textsuperscript{1}Univ. Rennes, INSA Rennes, CNRS, IETR - UMR 6164, Rennes, France.\\
\textsuperscript{2}KU 6G Research Center, Khalifa University, Abu Dhabi, UAE.\\
%Emails: firstname.lastname@insa-rennes.fr and firstname.lastname@ku.ac.ae
}
\maketitle
\begin{abstract}
This paper introduces a novel framework for designing efficient neural network architectures specifically tailored to \ac{tinyml} platforms. By leveraging \acp{llm} for \ac{nas}, a \ac{vit}-based \ac{kd} strategy, and an explainability module, the approach strikes an optimal balance between accuracy, computational efficiency, and memory usage. The \ac{llm}-guided search explores a hierarchical search space, refining candidate architectures through Pareto optimization based on accuracy, \acp{mac}, and memory metrics. The best-performing architectures are further fine-tuned using logits-based \ac{kd} with a pre-trained \ac{vit}-B/16 model, which enhances generalization without increasing model size. Evaluated on the CIFAR-100 dataset and deployed on an STM32H7 \ac{mcu}, the three proposed models, LMaNet-Elite, LMaNet-Core, and QwNet-Core, achieve accuracy scores of 74.50\%, 74.20\% and 73.00\%, respectively. All three models surpass current \ac{sota} models, such as MCUNet-in3/in4 (69.62\% / 72.86\%) and XiNet (72.27\%), while maintaining a low computational cost of less than 100 million \acp{mac} and adhering to the stringent 320 KB \ac{sram} constraint. These results demonstrate the efficiency and performance of the proposed framework for \ac{tinyml} platforms, underscoring the potential of combining \ac{llm}-driven search, Pareto optimization, \ac{kd}, and explainability to develop accurate, efficient, and interpretable models. This approach opens new possibilities in \ac{nas}, enabling the design of efficient architectures specifically suited for \ac{tinyml}. To facilitate further research and development in this field, the proposed framework and the best-performing architectures are made publicly available at \href{https://github.com/christophezei/llm-nas-kd-explainability}{Link}.
\end{abstract}

\acresetall
\section{Introduction}
\label{sec:introduction}
The proliferation of \ac{iot} devices built on tiny hardware platforms, such as \acp{mcu}, has underscored the demand for deploying \ac{dl} models in resource-constrained environments~\cite{ celzeinaty}. Despite their potential to democratize \ac{ai}, these devices face unique challenges that differ significantly from mobile \ac{dl}, particularly due to stringent memory limitations. For instance, the typical \ac{mcu} has a \ac{sram} of less than 512kB, which poses a substantial obstacle for running conventional \ac{dl} models~\cite{Lin_2023}.
\begin{figure}[t]
    \centering
\includegraphics[width=0.98\columnwidth]{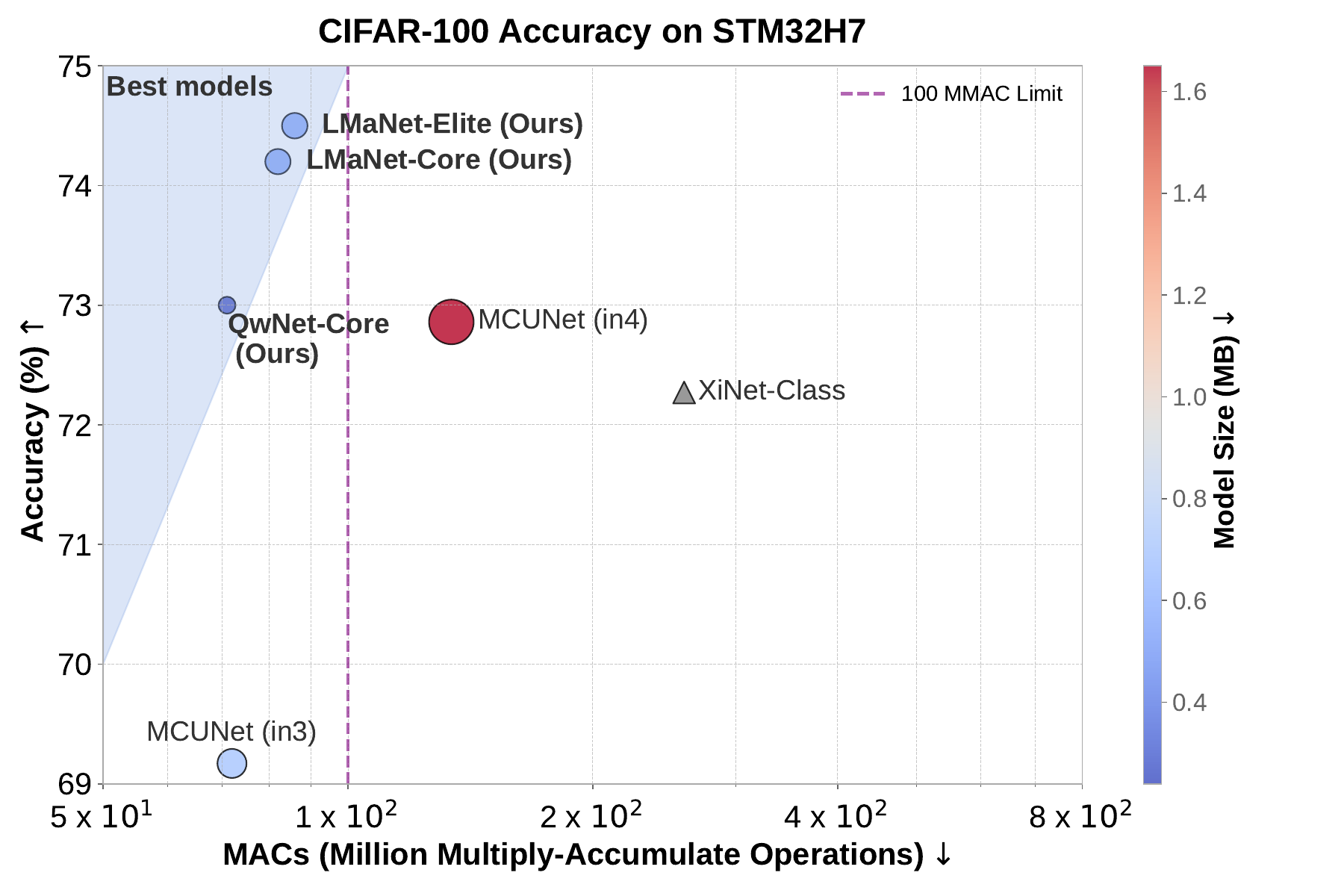}
    \caption{Our proposed architectures balance accuracy and efficiency, reducing model size and computational cost compared to baselines. Marker size and color indicate model size, XiNet-Class is shown as a triangle due to missing size information.}
    \label{fig:accuracy_vs_macs} \vspace{-6mm}
\end{figure}
Numerous lightweight \ac{dnn} architectures, such as MobileNet~\cite{Sandlermobilenets}, have demonstrated the utility of handcrafted model design for mobile applications. Yet, when adapted for \acp{mcu}, the limited memory resources pose significant hurdles. Even compact models often exceed the memory capacities of \acp{mcu}, requiring external memory or reliance on cloud-based solutions. To overcome these challenges, various optimization techniques, such as quantization, pruning, and \ac{kd}, have been extensively studied to reduce the size and complexity of neural networks~\cite{nagel2021white, NIPS1989_6c9882bb, hinton2015distilling}. A landmark contribution in this domain is the deep compression framework proposed by Han \textit{et al.}~\cite{han2016deep}, which combines pruning and quantization to significantly lower both memory usage and computational requirements. Recent works, such as MCUNet~\cite{mcunet, mcunetv2}, showcased advanced optimization strategies like patch-based inference and receptive field redistribution. These approaches significantly reduce peak memory usage, enabling deployment on tiny devices, but they rely on system co-design \ac{nas} with their custom inference engine. XiNet~\cite{Ancilotto2023XiNetEN}, on the other hand, adopts a manual design approach coupled with \ac{has} to develop efficient \ac{conv} architectures for \ac{tinyml}. Instead of relying on traditional \ac{nas}, XiNet prioritizes energy and memory efficiency through manual fine-tuning of architectural parameters. This approach demonstrates significant improvements in performance-energy trade-offs, achieving \ac{sota} results on platforms like the STM32H7. However, manual design inherently lacks the scalability and automation provided by modern \ac{nas} methodologies.
\begin{figure*}[t]
    \centering
    \includegraphics[width=\textwidth]{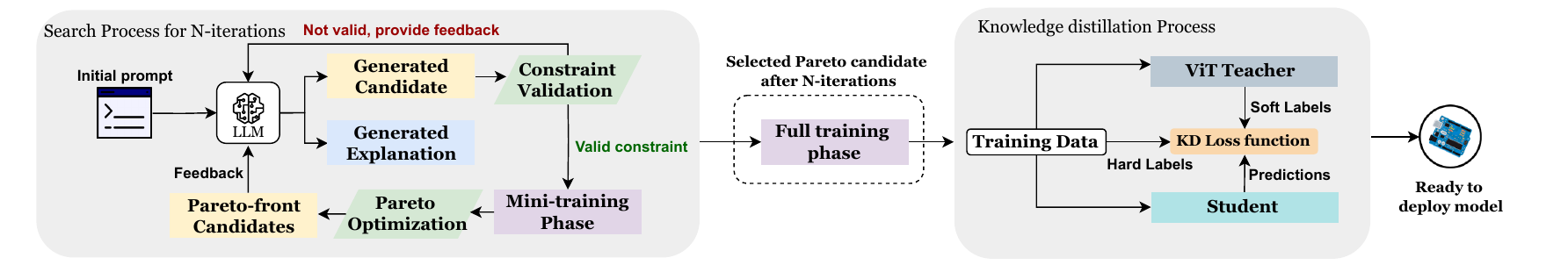} 
    %\caption{Overview of the LLM-guided architecture search with explainability. The LLM generates and validates candidate architectures based on MAC and SRAM constraints, followed by lightweight training and Pareto optimization. Explanations of design choices guide iterations, and the best candidates are refined using ViT-based knowledge distillation for the final model.}
    \caption{Overview of the LLM-guided architecture search process with integrated explainability. The LLM generates candidate architectures, which are validated against MAC and SRAM constraints. Valid candidates undergo lightweight training, followed by Pareto optimization to evaluate trade-offs between accuracy, computational cost, and memory usage. During the process, the LLM provides explanations for its design choices, enhancing interpretability and guiding further iterations. The best candidates are fully trained and refined using ViT-based knowledge distillation to produce the final deployable model.}
    \label{fig:full_process} \vspace{-4mm}
\end{figure*}

Recent advancements in \acp{llm} provide a new paradigm for architecture search~\cite{genius, nasir2023llmatic}. By leveraging their reasoning and generative capabilities, \acp{llm} streamline the design process, offering flexible and efficient solutions. However, these prior works primarily focus on general-purpose \ac{nas} tasks without specific deployment constraints. In this work, we bridge this gap by proposing a novel \ac{llm}-guided \ac{nas} framework tailored specifically for \ac{tinyml} platforms. Throughout this study, we aim to answer the research question of whether \acp{llm} can revolutionize the design of explainable and efficient \ac{tinyml} models. Our approach leverages open-source \acp{llm}, such as Llama~\cite{touvron2023llamaopenefficientfoundation} and Qwen~\cite{qwen}, to efficiently generate candidate architectures under stringent resource constraints. We integrate a Pareto-guided feedback loop to balance accuracy, computational cost, and memory footprint. Furthermore, our method incorporates \ac{kd}~\cite{hinton2015distilling}, using pretrained \ac{vit} models~\cite{wu2020visual} as teacher networks to enhance the performance of student architectures. As illustrated in Fig.~\ref{fig:accuracy_vs_macs}, our proposed architectures achieve \ac{sota} accuracy on the CIFAR-100 dataset~\cite{Krizhevsky2009LearningML} while maintaining low computational costs. In addition to optimizing neural architectures, our framework explores the explanatory potential of \acp{llm} for \ac{xai}, providing insights into the decision-making process during architecture generation. This aspect not only improves interpretability but also paves the way for more transparent and trustworthy deployment solutions in resource-constrained environments.
%In this work, we bridge this gap by proposing a novel \ac{llm}-guided \ac{nas} framework tailored specifically for \ac{tinyml} platforms.  Throuout this work, we attem to anwser the research question whether \acp{llm} can revolunize the design of explainable and efficient \ac{tinyml}. Our approach leverages open-source \acp{llm}, such as Llama~\cite{touvron2023llamaopenefficientfoundation} and Qwen~\cite{qwen}, to generate candidate architectures efficiently within stringent resource constraints. We integrate a pareto-guided feedback loop to balance accuracy, computational cost, and memory footprint. Furthermore, our method incorporates \ac{kd}~\cite{hinton2015distilling}, using pretrained \ac{vit} models~\cite{wu2020visual} as teacher networks to enhance the performance of student architectures. As illustrated in Fig.~\ref{fig:accuracy_vs_macs}, our proposed architectures achieve \ac{sota} accuracy on the CIFAR-100 dataset~\cite{Krizhevsky09learningmultiple} while maintaining low computational costs. In addition to optimizing neural architectures, our framework explores the explanatory potential of \acp{llm} for \ac{xai}, providing insights into the decision-making process during architecture generation. This aspect not only improves interpretability but also paves the way for more transparent and trustable deployment solutions in resource-constrained environments.
The contributions of this paper are as follows:
\begin{itemize}
    \item A novel \ac{llm}-guided \ac{nas} framework designed specifically for \ac{tinyml}, demonstrating significant reductions in search time compared to traditional methods.
    \item Pareto-guided optimization for balancing accuracy, computational cost, and memory usage, validated on the STM32H7 platform.
    \item An analysis of \ac{llm}-generated architectures, emphasizing the impact of model family and size on architectural efficiency.
    \item Exploration of \ac{llm}-driven explanations for \ac{xai}, enhancing transparency in architecture generation.
\end{itemize}

The remainder of this paper is as follows. Section~\ref{sec:related_work} provides an overview of related works. Section~\ref{sec:methodology} introduces the proposed framework, including its core components and workflow. Section~\ref{sec:experimental_setup} outlines the experimental results, followed by an ablation study in Section~\ref{sec:insights}, and a discussion of the \ac{xai} module in Section~\ref{sec:xai}. Finally, Section~\ref{sec:conclusion} presents the conclusions and future research directions.

\section{Related Works}
\label{sec:related_work}
\subsection{Advancements in Large Language Models}
The development of \acp{llm} has significantly advanced the field of \ac{ai}, enabling remarkable capabilities in tasks such as natural language processing, summarization, reasoning, and content generation~\cite{kaddour2023challengesapplicationslargelanguage}. Commercial models, including OpenAI's GPT series~\cite{openai2024gpt4technicalreport}, Google's Gemini~\cite{gemini2024gpt4technicalreport}, and Anthropic's Claude~\cite{TheC3}, have demonstrated the transformative potential of \acp{llm} across various domains, such as code generation~\cite{jiang2024surveylargelanguagemodels}, scientific discovery~\cite{ai4science2023impactlargelanguagemodels}, and interactive problem-solving. Concurrently, open-weight \acp{llm} like Llama~\cite{touvron2023llamaopenefficientfoundation}, Qwen~\cite{qwen}, Phi~\cite{abdin2024phi4technicalreport} and Falcon~\cite{almazrouei2023falconseriesopenlanguage} have emerged, offering accessible alternatives for researchers. These models provide the flexibility to explore \ac{llm} applications, fostering innovation and enabling greater transparency and control in the design and deployment of advanced \ac{ai} solutions. Building on these advancements, this study leverages open-weight \ac{llm} to guide \ac{nas}. Specifically, Llama3.2-3B-Instruct, Llama3.1-8B-Instruct~\cite{touvron2023llamaopenefficientfoundation} and Qwen2.5-3B-Instruct~\cite{qwen} models are employed to generate efficient neural architectures tailored for \ac{tinyml} platforms. Unlike traditional \ac{nas} methods, which often rely on computationally intensive strategies such as reinforcement learning or evolutionary algorithms~\cite{zoph2016neural, real2017large}, \acp{llm} enable a more rapid and flexible exploration of the design space through their inherent reasoning and generative capabilities~\cite{genius, nasir2023llmatic}.

\subsection{Evolution of Neural Architecture Search (NAS)}
\Ac{nas} represents a significant advancement in the automation of neural network design. Initial \ac{nas} approaches employed reinforcement learning to navigate vast search spaces~\cite{zoph2016neural}, while subsequent methods incorporated evolutionary algorithms~\cite{real2017large} and Bayesian optimization~\cite{bergstra2011algorithms} to refine the search process. However, the computational demands of these early methods were often prohibitive, leading to the development of more efficient strategies. For example, DARTS~\cite{liu2018darts} introduced a gradient-based approach, significantly streamlining the search process.  Further research has considerably expanded the scope of \ac{nas}, with a focus on techniques that balance computational efficiency with the generation of high-performing architectures~\cite{cai2018proxylessnas}. These advancements have facilitated the transition of \ac{nas} from theoretical exploration to practical deployment across a range of applications.
%\Ac{nas} represents a significant step forward in automating the design of neural network architectures. Initial \ac{nas} approaches relied on reinforcement learning to explore vast search spaces~\cite{zoph2016neural}, while later methods incorporated evolutionary algorithms~\cite{real2017large} and Bayesian optimization~\cite{bergstra2011algorithms} to refine the process. However, the computational demands of these early methods were often prohibitive, prompting the development of more efficient strategies. For instance, DARTS~\cite{liu2018darts} introduced a gradient-based approach to streamline the search process. Subsequent research has significantly expanded the landscape of \ac{nas}, focusing on methods that balance computational efficiency with high-quality architecture generation~\cite{cai2018proxylessnas}. These advances have enabled \ac{nas} to transition from theoretical exploration to practical deployment in a variety of applications.
\begin{figure*}[t]
    \centering
    \includegraphics[width=0.94\textwidth]{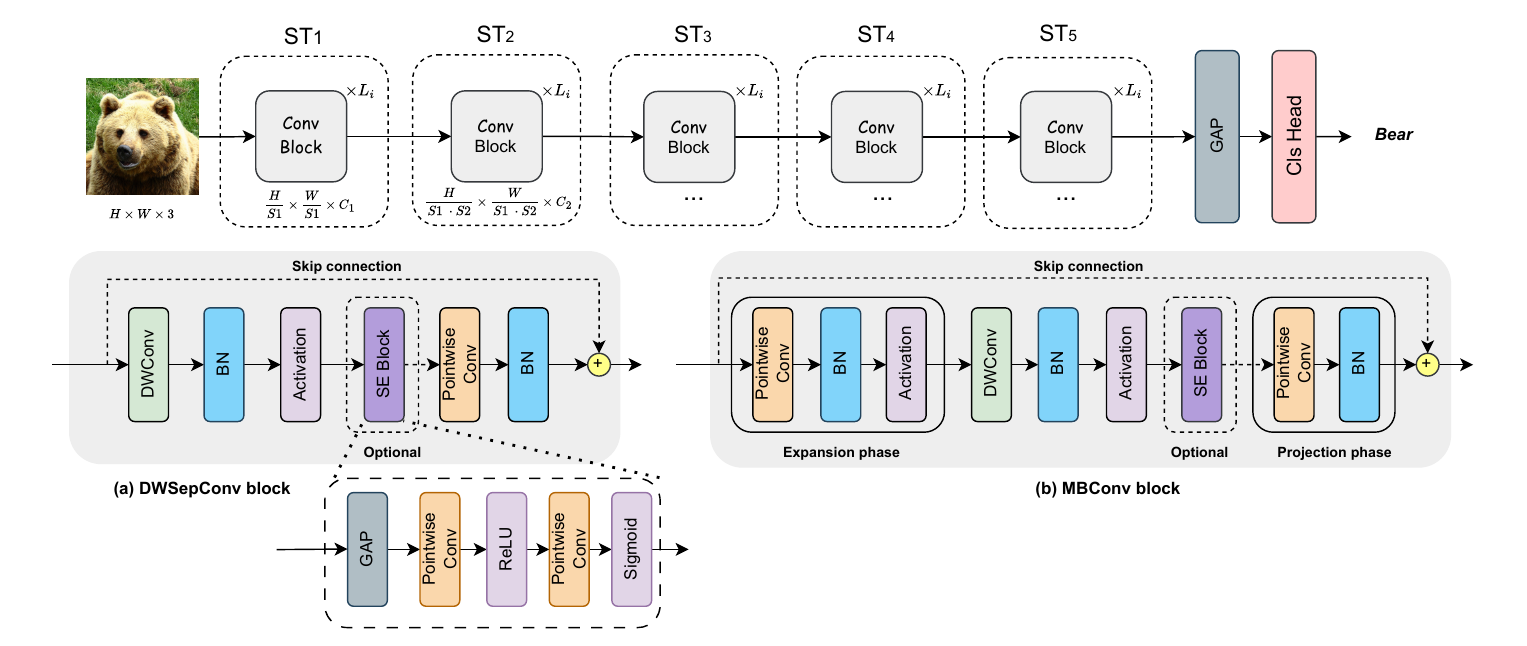}\vspace{-5mm}
    %\caption{Overview of the generalized hierarchical search space for designing lightweight and efficient DNN architectures in TinyML.}
    \caption{The \ac{llm} defines the configuration parameters for each stage, including layers \(L_i\), kernel size \(K\), stride \(S\), activation \(A\), convolution type \(Conv\ Block\), expansion \(E\), and output channels \(C_{\text{out}}\), along with the use of building blocks like \acs{dwsepconv} and \acs{mbconv}, and controls the inclusion of the \ac{se} block and skip connections.}
    \label{fig:general_search_space} \vspace{-4mm}
\end{figure*}
\subsection{LLMs for NAS: Current Approaches}
Recent advancements in leveraging \acp{llm} for \ac{nas} have demonstrated their potential to streamline the search process while maintaining competitive performance. GENIUS~\cite{genius} and LLMatic~\cite{nasir2023llmatic} are two notable studies exploring this synergy, each offering distinct methodologies but sharing certain limitations when applied to deployment-specific scenarios. GENIUS employs a prompting mechanism with GPT-4 to iteratively refine neural architectures. The process involves querying the \ac{llm} with architecture design tasks, followed by empirical evaluation and feedback integration to guide subsequent iterations. While GENIUS significantly reduces the computational cost compared to traditional \ac{nas} methods, its primary focus remains on optimizing for broad performance metrics such as accuracy. Moreover, the absence of a multi-objective framework and explainability limits its applicability in deployment-specific scenarios, particularly for resource-constrained environments. In contrast, LLMatic utilizes CodeGen-6.1B~\cite{Nijkamp2022CodeGenAO}, an \ac{llm} specialized in code generation, to generate and refine neural architectures. Starting with simple architectures, LLMatic applies \ac{qd} optimization techniques to explore a diverse set of solutions across different metrics, including parameter count and depth. This hierarchical evolution strategy ensures a variety of architectural candidates while maintaining a low computational footprint. However, LLMatic focuses more on architectural diversity than on deployment-specific constraints, and its reliance on code-generation \acp{llm} may limit its generalizability. Additionally, like GENIUS, it does not incorporate explainability into the architecture generation process.

These studies highlight the emerging role of \acp{llm} in automating and accelerating \ac{nas}. However, their limitations underscore the need for approaches that can address deployment-specific requirements, such as memory and computational constraints, while also enhancing transparency and interpretability.

\subsection{Bridging the Gap for TinyML Applications}
This work addresses the challenge by demonstrating the effectiveness of \acp{llm} in generating neural network architectures optimized for resource-constrained environments. Through an analysis of Pareto-optimal candidates, we validate the adaptability of \acp{llm} to such scenarios, highlighting their utility in tasks beyond general-purpose \ac{nas}. By integrating explainability into the search process, our approach not only bridges the gap between general-purpose and deployment-specific requirements but also enhances trust and usability through \ac{xai}. This framework underscores the potential of \acp{llm} to revolutionize \ac{nas} by making it faster, more efficient and interpretable, particularly for real-world deployments with stringent constraints.

%Our work addresses this challenge by demonstrating the efficacy of \acp{llm} in generating architectures optimized for resource-constrained environments. Through an analysis of pareto-optimal candidates, we validate the adaptability of \acp{llm} to such scenarios, highlighting their utility in tasks beyond general-purpose \ac{nas}. By integrating explainability into the search process, our approach not only bridges the gap between general-purpose and deployment-specific requirements but also enhances trust and usability through \ac{xai}. This framework underscores the potential of \acp{llm} to revolutionize \ac{nas} by making it faster, more efficient, and interpretable, particularly for real-world deployments with stringent constraints.

\section{Methodology}
\label{sec:methodology}
This section presents our novel approach to designing efficient \ac{tinyml} models by integrating \acp{llm} for \ac{nas} and \ac{vit}-based \ac{kd}. As shown in Fig.~\ref{fig:full_process}, the goal is to optimize architecture for computational efficiency and accuracy within \ac{tinyml} constraints. The method combines \ac{llm}-driven search, Pareto optimization~\cite{Deb2014}, and logits-based \ac{kd} with a pretrained \ac{vit}~\cite{wu2020visual}, while also incorporating an explainability module to provide insights into the \ac{llm}'s design choices, as discussed in Section~\ref{sec:xai}.
%This section presents our novel approach to design and deploy efficient \ac{tinyml} models by integrating \acp{llm} for \ac{nas} and \ac{vit}-based \ac{kd}. As illustrated in Fig.~\ref{fig:full_process}, the objective is to derive an optimal architecture that balances computational efficiency and predictive accuracy while adhering to the stringent constraints of \ac{tinyml} platforms. The approach leverages \ac{llm}-driven architecture search, Pareto optimization~\cite{Deb2014}, and logits-based \ac{kd} using a pretrained \ac{vit}~\cite{wu2020visual}. The framework also incorporates an explainability module, enabling the \ac{llm} to articulate its design choices. This module provides valuable insights into the \ac{llm}'s decision-making, as detailed in Section~\ref{sec:xai}.
\subsection{LLm-Guided Search Space Exploration}
\noindent \textbf{Search space.} The architecture search process is structured around a predefined skeleton composed of \(N\) sequential stages, as shown in Fig.~\ref{fig:general_search_space} for $N=5$. Each stage \(ST_i\) consists of a stack of \ac{conv} Blocks \(l_{i,j}\), where \(j \in \{1, \dots, L_i \}\), and all blocks \(l_{i,j}\) within a stage \(ST_i\) share the same configuration. This skeleton provides a consistent framework for generating lightweight and efficient \ac{dnn} architectures tailored to \ac{tinyml} platforms. The value of \(N\) defines the overall depth of the network and provides a consistent framework while enabling the \ac{llm} to optimize the architecture holistically. Unlike traditional approaches that configure each stage independently, this method allows the \ac{llm} to propose the entire \(N\) stage architecture in a single step. This global search approach ensures that the relationships and dependencies over the stages are considered, leading to more cohesive and contextually optimized architectures. At the core of each stage lies a generalized operation, repeated \(L_i\) times, defined as:
%\begin{equation}
%\mathbf{x}' \;=\;
%\Bigl(\textit{Conv block}(\mathbf{x})\Bigr)^{L_i}
%\label{eq:general_block_repeated}
%\end{equation}
\begin{equation}
\label{eq:general_block_repeated}
\mathbf{x}' \;=\; \mathrm{ConvBlock}_i(\mathbf{x})
\;=\;
\underbrace{
   f\bigl(f\bigl(\dots f(\mathbf{x}) \dots\bigr)\bigr)
}_{L_i \text{ times}},
\end{equation}
\noindent
where \( \mathbf{x} \) is the input tensor to the (i)-th stage, \( \mathbf{x}' \) is the resulting output tensor, \( f \) represents a single \ac{conv} block, and \( L_i \) indicates the number of \ac{conv} Blocks within this stage. The function \( f \) refers to the convolutional operation block, which can be either a \ac{dwsepconv} block (Fig.~\ref{fig:general_search_space}(a)) or a \ac{mbconv} block~\cite{Sandlermobilenets} (Fig.~\ref{fig:general_search_space}(b)), determining the computational structure of the block. To provide a more detailed analysis of the generalized formulation, each component of Eq.~\eqref{eq:general_block_repeated} is expanded and examined below based on its specific operation and role within the \ac{conv} Block.

 In the case of a \ac{dwsepconv} block, the convolution operation processes each input channel independently, significantly reducing computational cost while preserving spatial relationships. This is achieved through the \ac{dwconv} (\(\mathrm{dwconv}_{K \times K}\)), followed by batch normalization (\(\mathrm{BN}\)) and an activation function (\(A\)), such as ReLU6, Leaky ReLU, or Swish, which introduces non-linearity, enabling the model to capture complex patterns. The result is then passed through an optional \ac{se} block, denoted \(SEBlock(\mathbf{x})\), which recalibrates the importance of the channel by highlighting relevant characteristics at an intermediate step of the block. The inclusion of the \ac{se} block is controlled by the binary parameter \( \delta_{\text{se}} \), where \( \delta_{\text{se}} = 1 \) denotes the inclusion of the \ac{se} block within the \ac{conv} Block, and \( \delta_{\text{se}} = 0 \) denotes its absence. Additionally, the degree of channel reduction is governed by the \ac{se} ratio, selected from \(\{0.25, 0.5\}\), ensuring a balance between feature emphasis and computational efficiency. The intermediate tensor \(\mathbf{z}\) (prior to the \ac{se} block) is given by:
\begin{equation}
\mathbf{z} = A\Bigl( \mathrm{BN}( \mathrm{dwconv}_{K \times K}(\mathbf{x}) ) \Bigr) 
\label{eq:z_intermediate}
\end{equation}

The optional \ac{se} block computes the output tensor \({\tilde {\mathbf{z}}}\) as follows:
%If the \ac{se} block is included, the \ac{se} modified tensor \(\tilde{\mathbf{z}}\) is given by:
\begin{equation}
\tilde{\mathbf{z}} = \delta_{\text{se}} \, \bigl[ \mathbf{z} \odot \text{SEBlock}(\mathbf{z}) \bigr] + (1 - \delta_{\text{se}}) \, \mathbf{z}
\label{eq:se_module_inter}
\end{equation}

where the SEBlock operations for an input tensor \(\mathbf{z}\) is defined as:
\begin{equation}
 \text{SEBlock}(\mathbf{z}) = \mathbf{z} \odot \sigma \left( \text{Conv}_{1 \times 1} \left( \text{ReLU} \left( \text{Conv}_{1 \times 1} \left( \text{GAP}({\mathbf{z}}) \right) \right) \right) \right)
\label{eq:se_module}
\end{equation}

In this expression, \(\text{GAP}\) denotes \ac{gap}, \(\sigma\) is the sigmoid activation, \(\odot\) denotes element-wise multiplication, and \(\text{ReLU}\) is the activation function.

Subsequently, a \ac{pwconv} is applied, followed by batch normalization (\(\mathrm{BN}\)).  These operations adjust the number of channels and ensure the output tensor is ready for subsequent steps of the network. The resulting intermediate tensor after application of the \ac{pwconv} and batch normalization is defined as:
%Subsequently, a \ac{pwconv} is applied, followed by batch normalization (\(\mathrm{BN}_2\)). These operations adjust the number of channels and ensure the output tensor is ready for the next stage of the network. The intermediate tensor after applying the \ac{pwconv} and batch normalization is given by:
\begin{equation}
\mathbf{z}_{\mathrm{pw}} = \mathrm{BN}\Bigl( \mathrm{PWConv}(\tilde{\mathbf{z}}) \Bigr)
\label{eq:pw_intermediate}
\end{equation}

In the absence of a skip connection \( SC = 0 \), the output simplifies to \( \mathbf{x}' = \mathbf{z}_{\mathrm{pw}}\).
Conversely, when a skip connection is employed, it facilitates gradient flow and stabilizes training, particularly in deeper architectures. In this case, the resulting output, \( \mathbf{x}' \), is computed as follows:
\begin{equation}
\mathbf{x}' = \mathbf{x} + \mathbf{z}_{\mathrm{pw}}
\label{eq:skip_operation_dw}
\end{equation}
where \( \mathbf{x}\) is the input tensor and \(\mathbf{z}_{\mathrm{pw}}\) is the output tensor after the block's transformations.

Alternatively, when the \ac{conv} Block corresponds to a \acs{mbconv} block, the convolution operation follows a structure similar to that of the \acs{dwsepconv} block, but with an additional expansion phase at the beginning. This phase improves the model’s expressiveness while maintaining computational efficiency. Initially, the input tensor undergoes expansion through a \ac{pwconv}, which increases the number of channels by a factor \( E \), selected from \( \{3, 4, 6\} \). The expanded tensor is then passed through batch normalization \(BN\) and an activation function \(A\), as follows:

\begin{equation}
\mathbf{z} =  A \Bigl( \mathrm{BN} \Bigl( \mathrm{PWConv}(\mathbf{x}, E) \Bigr) \Bigr)
\label{eq:expansion_phase_mbconv}
\end{equation}

After the expansion, the tensor is processed through the same operations as in the \ac{dwsepconv} block, resulting in the intermediate tensor \(\mathbf{z}_{\mathrm{pw}}\), which is given by:

\begin{equation}
\mathbf{z}_{\mathrm{pw}} = \mathrm{DWSepConvBlock}(\mathbf{z})
\label{eq:z_intermediate_mbconv}
\end{equation}

It is important to note that although a \acs{dwsepconv} block is employed here, its internal skip connection is not utilized. Instead, the skip connection is implemented as defined for the \acs{mbconv} block. In the absence of a skip connection (\(SC = 0\)), the output simplifies to  \( \mathbf{x}' = \mathbf{z}_{\mathrm{pw}} \). Conversely, when a skip connection is active, the original input tensor \( \mathbf{x} \) is combined with the processed tensor \( \mathbf{z}_{\mathrm{pw}} \) using a residual connection, as detailed in Eq.~\eqref{eq:skip_operation_dw}.
\begin{figure}[t]
    \centering
    \resizebox{0.8\columnwidth}{!}{
    \begin{tcolorbox}[colframe=blue!50!black, colback=blue!5, sharp corners, boxrule=0.8pt, width=0.95\linewidth, title=Example LLM Prompt for Architecture Generation, fonttitle=\small]
    \small
    \textbf{Role:} You are a neural architecture design algorithm that exclusively outputs configurations in JSON format. \\[3pt]
    \textbf{Task:} Generate a lightweight neural network architecture tailored for CIFAR-100 image classification. \\[3pt]
    \textbf{Objective:} Achieve at least 70\% accuracy while minimizing computational cost and memory usage. \\[3pt]
    \textbf{Constraints:}
    \begin{itemize}
        \item Minimize RAM usage by reducing intermediate activation size.
        \item Prioritize stride=2 in early blocks for downsampling.
        \item Use smaller expansion\_factor and output\_channels in early blocks.
        \item Limit SE blocks and their ratios to reduce activation memory.
        \item Ensure total MACs $\leq$ 350M.
        \item Image size: 160$\times$160.
    \end{itemize}
    \textbf{Search Space:} Use only values from the hierarchical search space: json\_search\_space.
    \end{tcolorbox}
    }
    \caption{Example of initial prompt used to guide LLM architecture generation.}
    \label{fig:llm_prompt} \vspace{-4mm}
\end{figure}
Each \(\text{Conv Block}\) has searchable parameters selected by the \ac{llm}. These include the convolution kernel size \( K \) chosen from \( \{3, 5, 7 \} \), which defines the receptive field, the number of stacked blocks \( L_i \) selected from \( \{1, 2, 3, 4, 6\} \), which controls the stage's depth and representational capacity, the number of output channels \( C_{out} \) selected from \( \{16, 24, 32, 48, 64, 96, 128, 160\} \), which governs the stage's width and feature extraction capability, and the stride \( S \), set to either 1 or 2, which manages spatial downsampling.

As presented in Table~\ref{tab:search_space_parameters}, the search space for a single stage is further defined by the expansion factor \( E \), the presence of a \ac{se} block \(\delta_{\text{se}} \), the convolution block type (\ac{conv} Block), the presence of a skip connection, and the activation function $A$. These parameters (\(E, \, \delta_{\text{se}}, \,  \)  \ac{conv} Block, $SC, \, A$) along with $K$, $L_i$, $C_{out}$, and $S$ from the previous paragraph, offer 3, 2, 2, 2, 3, 3, 5, 8, and 2 choices, respectively, yielding approximately 17,280 unique configurations per stage. All blocks within a stage share the same configuration. Thus, $L_i$ acts as a discrete parameter in the search space rather than increasing individual stage complexity. Empirical evaluations indicated that a skeleton architecture with $N = 5$ stages provides an optimal balance between accuracy and efficiency, making it well-suited for resource-constrained environments. This configuration ensures the generated architectures meet stringent memory and computational constraints while maintaining competitive performance. Consequently, with $N = 5$ independently configured stages, the total search space encompasses approximately $17,280^5$ possible architectures.\\\\
\textbf{Search algorithm.} The proposed search algorithm leverages the capabilities of a pre-trained \ac{llm} to explore the search space described earlier effectively. Instead of fine-tuning, the \ac{llm} is guided through carefully crafted prompts designed to encode architectural constraints and optimization objectives. These prompts allow the \ac{llm} to generate candidate architectures that adhere to the stringent requirements of \ac{tinyml} platforms, such as constraints on memory, the maximum allowed \acp{mac}, and peak \ac{sram} usage. An example of the prompt is provided in Fig.~\ref{fig:llm_prompt}. To further improve efficiency, a series of validation checks is applied to each generated candidate before training. First, the estimated \acp{mac} of the architecture must lie within the defined range,  M\acp{mac}\_range = [min, max]. If a candidate exceeds or falls below these thresholds, it is rejected, and feedback is provided to the \ac{llm} through the prompts, specifying the current value and the desired range. Similarly, the architecture's estimated peak \ac{sram} must not exceed a threshold for a quantized \texttt{int8} model. Candidates that violate this constraint are also rejected, with similar feedback provided to the \ac{llm}. Additionally, to avoid redundant computations, architectures that have already been generated in previous iterations are skipped, ensuring only unique candidates proceed to the training phase. By applying these checks, the search algorithm significantly reduces the time and resources spent training invalid or redundant candidates. Only architectures that pass all validation checks are considered legitimate for training, forming the basis for subsequent refinement through iterative feedback.
\begin{table}[t]
\centering
\caption{Configuration options for the architecture search space.}
\resizebox{\columnwidth}{!}{%
\begin{tabular}{llc}
\toprule
\textbf{Parameter}         & \textbf{Description}                              & \textbf{Choices} \\
\midrule
\(C_{\text{out}}\)         & Output channels                                   & \{16, 24, 32, 48, 64, 96, 128, 160\}\\
\(K\)                      & Kernel size                                      & \{3, 5, 7\} \\
\(S\)                      & Stride                                           & \{1, 2\} \\
\(E\)                      & Expansion factor                                 & \{3, 4, 6\}\\
\(\delta_{\text{se}}\)                     & Enable the \ac{se} block          & \{True, False\} \\
\( \text{Conv Block}\)                     & Convolution type                                 & \{\acs{dwconv}, \ac{mbconv}\} \\
\(SC\)                     & Skip connection                                   & \{True, False\}\\
\(A\)                      & Activation function                              & \{ReLU6, Leaky ReLU, Swish\}\\
\(L_i\)                    & Number of layers within the block                & \{1, 2, 3, 4, 6\}\\
\midrule
\textbf{Total per stage}   & \multicolumn{2}{c}{\(\approx 17,280\) configurations} \\
\bottomrule
\end{tabular}%
}
\label{tab:search_space_parameters} \vspace{-4mm}
\end{table}
\subsection{Feedback Loop With Pareto Optimization}
To iteratively refine the architectures generated by the \ac{llm}, we implement a feedback loop guided by pareto optimization. Each valid candidate architecture proposed by the \ac{llm} undergoes a mini-training phase to evaluate its performance. Specifically, the candidate is trained on the CIFAR-100~\cite{Krizhevsky2009LearningML} dataset for 30 epochs, providing an efficient yet consistent estimation of the key metrics required for feedback. These metrics include test accuracy, the number of \acp{mac}, and the total number of model parameters. Mini-training ensures that meaningful performance feedback is obtained without incurring excessive computational overhead, enabling a fast and iterative refinement process.
Once the performance metrics are obtained, pareto optimization is applied to analyze the trade-offs between the competing objectives. Pareto optimization is a multi-objective optimization approach that identifies solutions achieving optimal trade-offs, where improving one objective would lead to the degradation of another. A solution is considered \textit{Pareto optimal} if no other solution in the search space outperforms it across all metrics simultaneously. The collection of such solutions forms the \textit{Pareto front}, representing the set of non-dominated candidates that balance the conflicting objectives.

In this work, we evaluate candidate architectures using three critical metrics: test accuracy, the number of \acp{mac}, and the total number of parameters. Accuracy reflects the classification performance on the test dataset. \Acp{mac} quantify computational cost, while the total number of parameters influences the model’s memory footprint, a key constraint for deployment on resource-limited devices. These objectives are inherently conflicting: improving accuracy may require increasing model complexity, while reducing computational cost and memory often comes at the expense of performance.

To navigate these trade-offs, we employ pareto dominance as the decision criterion. A solution \( A \) is said to dominate a solution \( B \) if it performs no worse than \( B \) across all metrics and is strictly better on at least one metric. While dominance suggests eliminating clearly inferior solutions, pareto optimization deliberately retains all non-dominated candidates to ensure that trade-offs between accuracy, computational cost, and memory usage are fully explored. For instance, one candidate might achieve superior accuracy but at the cost of higher \acp{mac}, while another may offer a more resource-efficient configuration with slightly lower accuracy. By maintaining the pareto front, we preserve this diversity, ensuring flexibility when selecting architectures for different deployment constraints. The pareto front is dynamically updated as new candidates are evaluated. Each new architecture is compared to the current pareto front: solutions dominated by the new candidate are removed, while non-dominated candidates are retained. If the new architecture itself is non-dominated, it is added to the pareto front. This ensures that only candidates representing optimal trade-offs remain. Additionally, we track the single architecture achieving the highest test accuracy, even if it does not meet resource constraints. This ensures that the architecture with the best overall performance is preserved, complementing the pareto-optimal solutions.

The updated pareto front and the best accuracy candidate are then used to generate structured feedback for the \ac{llm}. This feedback provides a summary of the current set of high-quality candidates, including their accuracy, \acp{mac}, and total number of parameters. Additionally, trends across the pareto front, such as the average accuracy or computational cost, are analyzed to guide the search process. Targeted suggestions are included to refine future candidates. For example, the feedback may encourage reducing \acp{mac} or parameters in solutions that exceed predefined thresholds, or it may focus on improving accuracy while maintaining resource efficiency.

By iteratively evaluating, filtering, and refining candidate architectures, the feedback loop enables the \ac{llm} to progressively converge toward solutions that achieve an optimal balance between predictive performance and computational efficiency. Once a candidate demonstrates strong performance in the pareto front or achieves the best accuracy, it is selected for full training. In the full training phase, the architecture is trained for an extended number of epochs to fully exploit its learning capacity. This final model is subsequently passed to the \ac{vit}-based \ac{kd} phase, where additional refinements are applied to enhance accuracy further without increasing the model size or computational overhead.
\subsection{ViT-Based Knowledge Distillation (KD)}
\Ac{kd} is a model compression technique where a smaller student model learns to approximate the behavior of a larger teacher model. In this work, we employ logits-based \ac{kd}, which focuses on aligning the output probabilities of the student and teacher models. This approach is particularly suitable for resource-constrained platforms, as it enhances accuracy without increasing model size or computational cost. Logits-based \ac{kd} leverages softened output probability distributions to transfer knowledge from the teacher to the student. Given teacher logits \( z_t \) and student logits \( z_s \), the softened probabilities are defined as:
\begin{equation}
p_t = \text{softmax}\left(\frac{z_t}{T}\right), \quad p_s = \text{softmax}\left(\frac{z_s}{T}\right),
\end{equation}
where \( T > 1 \) is the temperature parameter. A higher \( T \) produces smoother distributions, emphasizing the relative probabilities of non-target classes. This softening enables the student model to capture inter-class relationships, improving generalization.

The distillation loss \( \mathcal{L}_{\text{KD}} \) is expressed as the \ac{kl} divergence~\cite{Perez} between the teacher and student probabilities:
\begin{equation}
\mathcal{L}_{\text{KD}} = T^2 \cdot \text{KL}\left(p_t \parallel p_s\right) = T^2 \sum_{i=1}^C p_t^{(i)} \log\left(\frac{p_t^{(i)}}{p_s^{(i)}}\right),
\label{eq:kd_loss}
\end{equation}
where \( C \) is the total number of classes. The scaling factor \( T^2 \) ensures appropriate gradient magnitudes during optimization. The overall loss function combines the \ac{kd} loss with the cross-entropy loss \( \mathcal{L}_{\text{CE}} \), based on ground-truth labels \( y \):
\begin{equation}
\mathcal{L} = \alpha \, \mathcal{L}_{\text{CE}} + (1 - \alpha) \, \mathcal{L}_{\text{KD}},
\label{eq:total_loss}
\end{equation}
where \( \alpha \) controls the relative contribution of each loss component. To ensure an effective balance between ground-truth supervision and distillation, we adopt an adaptive scheduling strategy for \( \alpha \). Initially, the student model relies more heavily on the cross-entropy loss to establish a strong understanding of the task. As training progresses, the influence of the distillation loss increases gradually. This is achieved by updating \( \alpha \) at each epoch using:
\begin{equation}
\alpha = \alpha + (\alpha_{\text{final}} - \alpha)  \left(\frac{\text{epoch}}{\text{num\_epochs}}\right),
\label{eq:alpha_update}
\end{equation}
where \( \alpha_{\text{final}} \) is the desired final value. This gradual transition avoids overwhelming the student with complex teacher knowledge early in training and ensures that it benefits fully from the softened logits in later stages. We set a high temperature in our experiments, as higher values effectively capture fine-grained class relationships. For instance, softened logits can encode similarities between visually similar classes, such as “cat” and “tiger,” which are otherwise challenging for the student to learn from hard labels alone.

The \ac{vit}~\cite{wu2020visual} is employed as the teacher model in this work. Unlike \acp{cnn} that focus on local spatial features, \acp{vit} use self-attention mechanisms to capture global contextual relationships across the input image. This allows the teacher to produce enriched output probabilities that better reflect subtle relationships between classes. The student model benefits from this knowledge transfer by learning a richer representation of the dataset, leading to improved generalization.
\section{Experimental Results}
\label{sec:experimental_setup}
\subsection{Experimental setup}
\subsubsection{LLM Models} 

The neural \ac{nas} process is driven by open-weight \acp{llm} to ensure accessibility and reduce local execution costs. Specifically, Llama3.2-3B-Instruct and Llama3.1-8B-Instruct~\cite{touvron2023llamaopenefficientfoundation}, along with Qwen2.5-3B-Instruct~\cite{qwen}, are employed for their efficiency, with parameter sizes up to 8 billion. This parameter range represents a suitable balance between performance and computational cost. The use of their instruction-tuned versions ensures more coherent outputs. Furthermore, leveraging open-source models enhances the transparency and reproducibility of the \ac{nas} workflow.%The \ac{nas} process is driven by open-weight \acp{llm} to ensure accessibility and lower local execution costs. Llama3.2-3B-Instruct and Llama3.1-8B-Instruct~\cite{touvron2023llamaopenefficientfoundation}, as well as Qwen2.5-3B-Instruct~\cite{qwen}, are selected for their efficiency, with sizes up to 8B parameters, a reasonable range for balancing performance and computational cost. Their instruct-tuned versions ensure coherent outputs. The use of open-source models also supports transparency and reproducibility in the workflow.
%The \ac{nas} process is driven by open-source \acp{llm}, ensuring accessibility and reproducibility. For this purpose, we utilize two versions of the Llama model, specifically \textbf{Llama3.2-3B-Instruct} and \textbf{Llama3.1-8B-Instruct}~\cite{touvron2023llamaopenefficientfoundation}. The instruct-tuned versions were chosen because they are well-suited for task-specific prompts. The use of two model sizes enables us to evaluate the impact of model scale on the quality of generated architectures: the smaller 3B version ensures computational efficiency during training, while the larger 8B version tests whether increasing the \ac{llm}'s parameters leads to improved architecture proposals. To validate the generalizability of our approach, we further include the \textbf{Qwen2.5-3B-Instruct} model~\cite{qwen}, demonstrating that the proposed method is not restricted to a single \ac{llm} family. The choice of open-source models reflects our emphasis on transparency and ensures that the workflow remains accessible for further research and deployment.
\subsubsection{Dataset}
The CIFAR-100 dataset~\cite{Krizhevsky2009LearningML} is employed to evaluate the generated architectures. This dataset comprises 60,000 images distributed across 100 classes, with a training set of 50,000 images and a test set of 10,000 images. For the experiments, images are resized to 160$\times$160 pixels and standard preprocessing techniques, including normalization, are applied. CIFAR-100 was chosen for its balanced class distribution, which allows the evaluation to focus on the representational capacity of the generated backbone architectures.
%The CIFAR-100 dataset~\cite{Krizhevsky09learningmultiple} is used to evaluate the generated architectures. It contains 60,000 images across 100 classes, split into 50,000 for training and 10,000 for testing. Images are resized to \( 160 \times 160 \), with standard preprocessing including normalization. CIFAR-100 is selected for its balanced class count, which allows focus on evaluating the backbone's representational capacity.
%The CIFAR-100 dataset~\cite{Krizhevsky09learningmultiple} is used to evaluate the generated architectures. It contains 60,000 images across 100 classes, split into 50,000 for training and 10,000 for testing. Images are resized to \( 160 \times 160 \), with standard preprocessing steps including normalization. AutoAugment~\cite{Dogus} is applied to increase data diversity through policy-based transformations. While ImageNet~\cite{imagenet} improves generalization via transfer learning, we opted for CIFAR-100 to emphasize the backbone's representational capacity. ImageNet's large class count can shift focus to the classification head, masking the backbone's feature learning.

\subsubsection{Training Details}
The training process is structured into three distinct phases: a lightweight mini-phase for rapid initial evaluation of candidate architectures, a full training phase to optimize the final selected model, and a \ac{kd} phase for further fine-tuning. Each phase is tailored with specific data augmentations and hyperparameter settings to balance computational efficiency and model performance. \\\\
\textbf{Mini-Phase Training.}
In this initial phase, each candidate architecture generated by the \ac{llm} undergoes a lightweight evaluation to provide rapid feedback for Pareto optimization while minimizing computational overhead. The candidates are trained for 30 epochs on the CIFAR-100 dataset with a batch size of 128. Training begins with a learning rate of 0.5, which is progressively reduced by a factor of 0.1 every 10 epochs using a step learning rate scheduler. A warm-up period of 10 epochs is employed to ensure gradient stability during the initial stages of training. The \ac{sgd} optimizer is used with Nesterov momentum set to 0.9 and weight decay set to \( 10^{-4} \). Standard data augmentations, including image resizing to 160$\times$160 pixels and AutoAugment~\cite{Dogus}, are applied to improve generalization. Advanced augmentations, such as MixUp, are excluded to avoid introducing unnecessary complexity during the limited 30-epoch training phase. This streamlined setup ensures both a fair comparison and efficient convergence across candidate architectures. \\\\
%In this phase, each candidate architecture generated by the \ac{llm} undergoes lightweight evaluation to provide fast feedback for pareto optimization while minimizing computational overhead. The candidates are trained for 30 epochs on the CIFAR-100 dataset with a batch size of \( 128 \). Training begins with a learning rate of \( 0.5 \), which is progressively reduced by a factor of \( 0.1 \) every 10 epochs using a step learning rate scheduler. A warm-up period of \( 10 \) epochs ensures gradient stability during the early stages of training. We use the \ac{sgd} optimizer with Nesterov momentum (\( 0.9 \)) and weight decay set to \( 10^{-4} \). Standard augmentations, including image resizing to \( 160 \times 160 \) and AutoAugment~\cite{Dogus}, are applied to improve generalization. Advanced augmentations, such as MixUp, are excluded to avoid unnecessary difficulty during the limited 30-epoch training phase. This streamlined setup ensures both fair comparison and efficient convergence across candidate architectures.\\
\textbf{Full-Phase Training.} The architecture selected from the Pareto front undergoes full training to optimize its performance. During this phase, the model is trained for 120 epochs with a batch size of 128, using the \ac{sgd} optimizer with Nesterov momentum set to 0.9 and weight decay set to \( 10^{-4} \). The learning rate follows a warmup-cosine annealing schedule: it begins at 0.0, increases linearly to 0.5 during the first 20 epochs (the warmup phase), and then decays smoothly following a cosine function for the remainder of the training. To improve generalization, AutoAugment~\cite{Dogus} is employed alongside MixUp~\cite{Zhang} with an alpha value of 0.2. These augmentations enhance robustness while preventing overfitting, ensuring that the final model achieves optimal performance. \\\\
{\bf \ac{kd} Phase Training.} In this final fine-tuning phase, logits-based \ac{kd} is applied to further refine the model. A pretrained \ac{vit}-B/16\footnote{\url{https://huggingface.co/google/vit-base-patch16-224-in21k}} serves as the teacher model, transferring its knowledge to the student model (the selected architecture). The distillation process uses a temperature \( T = 10 \) to soften the teacher's outputs, allowing the student to learn inter-class relationships. The total loss combines cross-entropy and distillation losses, weighted by an adaptive alpha scheduling strategy (\( \alpha = 0.4 \), \( \alpha_{\text{final}} = 0.8 \)) as defined in~\eqref{eq:alpha_update}. Since this phase focuses on fine-tuning rather than extensive retraining, no advanced data augmentations, such as MixUp or AutoAugment, are applied. Only basic preprocessing, including image resizing to 160$\times$160 pixels and normalization, is performed to maintain consistency with earlier phases. The \ac{kd} phase runs for 50 epochs, allowing the student to refine its performance efficiently while adhering to computational constraints.
\subsubsection{Hardware}
The models are trained on an NVIDIA RTX-8000 GPU server equipped with 4608 CUDA cores and 48 GB of VRAM, running a Linux-based operating system. For inference evaluation, the models are profiled on an STM32H743 \ac{mcu}, which features 2 MB of flash memory and 512 KB of \ac{sram}. Although the STM32H743 has 512 KB of \ac{sram}, the models are specifically tested to meet the more stringent constraint of 320 KB \ac{sram}, a common limitation in many \ac{tinyml} applications. Furthermore, the minimum and maximum M\acp{mac} are set to 70 and 350, respectively, to ensure the models operate within the defined computational limits. This setup reflects the real-world constraints of \ac{tinyml} platforms, ensuring that the designed architectures are both efficient and suitable for deployment in resource-constrained environments.
%The models are trained on an NVIDIA RTX-8000 GPU server equipped with 4608 CUDA cores and 48 GB of VRAM, running a Linux-based operating system. For inference evaluation, the models are profiled on an STM32H743 \ac{mcu}, which features 2 MB of flash memory and 512 KB of \ac{sram}. Although the STM32H743 has 512 KB of \ac{sram}, the models are specifically tested to meet the more stringent constraint of 320 KB \ac{sram}, a common limitation in many \ac{tinyml} applications. This setup reflects the real-world constraints of \ac{tinyml} platforms, ensuring that the designed architectures are both efficient and suitable for deployment in resource-constrained environments.
\subsubsection{Baselines and Evaluation Metrics}
The effectiveness of the proposed method is demonstrated by comparing our optimized architectures LMaNet-Core, QwNet-Core, and LMaNet-Elite which are generated using Llama3.2-3B-Instruct, Qwen2.5-3B-Instruct, and Llama3.1-8B-Instruct, respectively, against \ac{sota} baselines for resource-constrained platforms. These baselines include lightweight models such as XiNet~\cite{Ancilotto2023XiNetEN} and various versions of MCUNet~\cite{mcunet}, known for their balance of efficiency and accuracy. Model performance is evaluated using four key metrics: classification accuracy on the CIFAR-100 test dataset, number of \acp{mac} to assess computational complexity, total model parameters to evaluate memory usage, and peak \ac{sram} usage during inference to ensure hardware compatibility. These metrics provide a comprehensive assessment of accuracy, computational cost, and resource efficiency.
\begin{table}[t]
\centering
\caption{Comparison of CIFAR-100 classification performance with \ac{sota} baselines for \(160 \times 160\) input resolution.}
\label{tab:baseline_comparison}
\resizebox{\columnwidth}{!}{
\begin{tabular}{@{}l>{\columncolor{lightblueAlpha}}c>{\columncolor{lightblueAlpha}}c>{\columncolor{lightblueAlpha}}c>{\columncolor{white}}c>{\columncolor{white}}c@{}}
\toprule
\textbf{Models}           & \textbf{Accuracy}       & \textbf{MACs}             & \textbf{Flash}            & \textbf{\#Parameters }       & \textbf{Search cost} \\
           & \textbf{(\%)$\uparrow$}       & \textbf{(M)$\downarrow$}             & \textbf{(MB)$\downarrow$}            & \textbf{(M)$\downarrow$}       & \textbf{(days)}$\downarrow$ \\ \midrule
MCUNet-in4~\cite{mcunetv2} (NeurIPS21)   & 72.86                      & 134                              & 1.65                                   &                      1.73          &      12.5                       \\
MCUNet-in3~\cite{mcunetv2} (NeurIPS21)   & 69.62                         & 72                               & 0.70                                   &                 0.74               &      12.5                       \\
XiNet-Class~\cite{Ancilotto2023XiNetEN} (ICCV23) & 72.27                        & 259                              & --                                     &                         --         &      manual                       \\
\midrule
 LMaNet-Core (Ours)     & 74.20                         & 82                              & 0.53                                  &   0.51                             &                2.5             \\
QwNet-Core (Ours)    & 73.00                        &     \textbf{71}                        & \textbf{0.24}                          &     \textbf{0.18}                           &      3.5                       \\
LMaNet-Elite (Ours)    & \textbf{74.50}                & 86                               & 0.54                                   &                      0.44          &       \textbf{1.5}                     \\
\bottomrule
\end{tabular}%
}\vspace{-4mm}
\end{table}
\subsection{Results and Analysis}
\label{sec:results}
\subsubsection{Accuracy, Computational Complexity, and Model Size Trade-offs} 
Table~\ref{tab:baseline_comparison} presents a detailed comparison of our proposed models with \ac{sota} baselines in terms of accuracy on the CIFAR-100 dataset, \acp{mac}, model size and search cost. Among the proposed models, LMaNet-Elite achieves the highest accuracy at 74.50\%, surpassing the MCUNet-in4 model by 1.64\% and XiNet-Class by 2.23\%. LMaNet-Elite also demonstrates superior efficiency, with a computational cost of 86 million \acp{mac}, significantly lower than MCUNet-in4 with 134 million \acp{mac}. Furthermore, LMaNet maintains a compact model size of 0.54 MB, making it an ideal choice for \ac{tinyml} platforms. While MCUNet-in3 and XiNet-Class models also achieve competitive accuracy, they exhibit trade-offs. MCUNet-in3 achieves a lower accuracy of 69.62\% with a computational cost of 72 million \acp{mac}, whereas XiNet-Class incurs significantly higher computational costs (259 million \acp{mac}), making it less efficient than LMaNet-Elite despite its competitive accuracy of 72.27\%.
LMaNet-Core and QwNet-Core also surpass the \ac{sota} models, achieving competitive accuracies of 74.20\% and 73.00\%, respectively, while using smaller model sizes of 0.51MB (LMaNet-Core) and 0.24MB (QwNet-Core). These models strike an optimal balance between performance and memory efficiency, with QwNet-Core being especially suitable for highly resource-constrained devices due to its compact model size of 0.24MB. In addition, our models significantly outperform MCUNet in search cost. LMaNet-Elite, LMaNet-Core, and QwNet-Core require only 1.5, 2.5, and 3.5 days, respectively, over 4 times faster than MCUNet’s 12.5 days. This demonstrates the efficiency of the \ac{llm}-guided methodology in reducing search time while adhering to the stringent resource constraints of the STM32H743 \ac{mcu}.
\subsubsection{SRAM Usage and Memory Efficiency}
\begin{figure}[t]
    \centering
    \includegraphics[width=0.9\columnwidth]{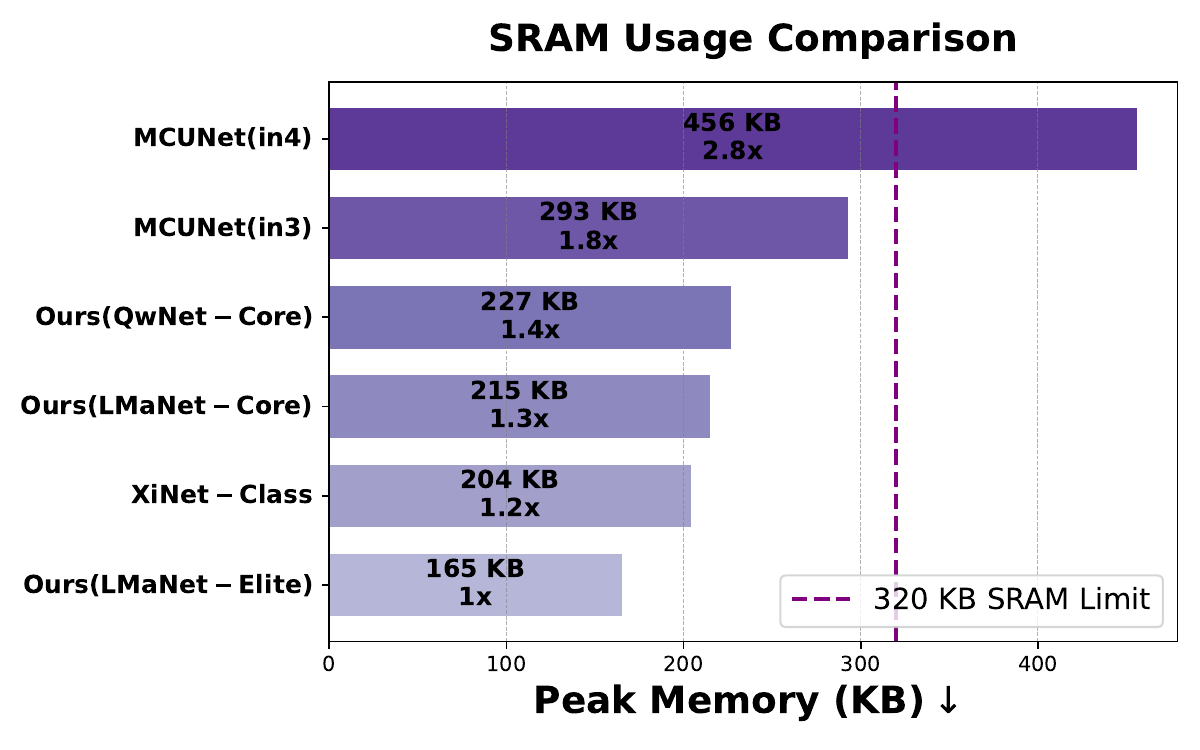}
    \caption{SRAM usage comparison of different models for deployment on STM32H7. The purple dashed line indicates the 320 KB SRAM constraint. Each bar shows the SRAM usage (in KB) of the models, with comparative factors relative to \textbf{LMaNet-Elite}, which achieves the lowest peak memory consumption.}
\label{fig:sram_usage_comparison} \vspace{-4mm}
\end{figure}

Figure~\ref{fig:sram_usage_comparison} provides a visual comparison of the \ac{sram} usage across different models, emphasizing their suitability for deployment in memory-constrained environments. The LMaNet-Elite model shows the lowest \ac{sram} usage at 165KB, well below the 320KB limit, making it ideal for highly resource-constrained devices. In contrast, the MCUNet-in4 model, with a peak \ac{sram} usage of 456KB, exceeds the 320KB limit, limiting its deployment to systems with higher memory capacities. The LMaNet-Core and QwNet-Core models also fit within the 320KB \ac{sram} limit, with usages of 215KB and 227KB, respectively. These models offer a strong balance between memory efficiency and performance, making them suitable for environments with stringent memory constraints. Notably, all models fit comfortably under the more relaxed 512KB \ac{sram} limit, ensuring compatibility with platforms like the STM32H743 \ac{mcu}. While both MCUNet-in3 and XiNet-Class meet the 320KB \ac{sram} limit, they come with trade-offs in other areas as discussed earlier. MCUNet-in3 uses 293KB of \ac{sram} but sacrifices accuracy. On the other hand, XiNet-Class, despite using only 204KB of \ac{sram}, incurs significantly higher computational costs, making it less efficient than our proposed models.
These results demonstrate the strength of our \ac{llm}-guided framework in generating models that balance high accuracy with low computational cost and minimal memory usage, pushing the boundaries of \ac{tinyml} for real-world applications.
\section{Ablation Study}
\label{sec:insights}
This ablation study investigates the impact of search efficiency with \acp{llm}, focusing specifically on the performance of Llama3.2-3B-Instruct, Llama3.1-8B-Instruct, and Qwen2.5-3B-Instruct. Furthermore, the study explores the influence of the proposed \ac{vit}-\ac{kd} method in further improving model accuracy, examining how it enhances the performance achieved through the configuration search phase.
\subsection{Search efficiency with \acp{llm}}  In the experiments, \ac{sram} is treated as a hard constraint, while accuracy, \acp{mac}, and the number of parameters serve as soft constraints optimized by the \ac{llm} during the search process to update the Pareto front. To ensure a fair comparison among the three \acp{llm}, the same training configuration is used across the mini-phase training, full training phase, and \ac{vit}-\ac{kd} phase. Regarding the \ac{llm} configuration parameters, the decoding temperature is set to \( T = 1.5 \) and the minimum probability threshold (\texttt{min\_p}) is set to \( 0.1 \). A higher temperature promotes exploration by increasing randomness in the output, which helps the \ac{llm} generate diverse configurations. Simultaneously, the \texttt{min\_p} threshold prevents the selection of excessively low-probability tokens, thereby reducing the likelihood of incoherent or invalid outputs. This balance between exploration and control is crucial for effectively navigating the hierarchical search space while adhering to task constraints. By fixing these variables and maintaining the same configuration, the choice of \ac{llm} becomes the sole factor in the experiment. The experiment is conducted over 500 iterations, where all three models adhere to the \ac{sram} hard constraint. However, significant differences are observed in the time spent on the search, the final model accuracy, and \ac{sram} usage. Table~\ref{tab:model_comparison} presents several key observations regarding the efficiency and effectiveness of the different \acp{llm} in identifying suitable configurations. For each model, the total number of generated candidates, the number of Pareto candidates, and the average values for accuracy, \acp{mac}, and number of parameters of the Pareto candidates are reported. Llama 3B generated a total of 127 candidates, of which 27 qualified as Pareto candidates. The average accuracy of the Pareto candidates was 60.50\%, with an average of 112.01 million \acp{mac} and 0.15 million parameters. Despite having a smaller search space relative to the other models,  Llama 3B still identified viable configurations that satisfied the \ac{sram} constraint, demonstrating its efficiency in terms of both search time and candidate generation. The search process for this model required 2.5 days as indicated in Table~\ref{tab:baseline_comparison}. Llama 8B produced 100 total candidates, with 29 Pareto candidates. 
\begin{table}[t]
\centering
\caption{Performance metrics for model generation across \ac{llm} models in the mini-phase (30 epochs), presented in the format [Min, Max] Avg. TC: total candidates, PC: Pareto candidates.}
\resizebox{\columnwidth}{!}{
\begin{tabular}{@{}l>{\columncolor{white}}c>{\columncolor{white}}c>{\columncolor{lightblueAlpha}}c>{\columncolor{lightblueAlpha}}c>{\columncolor{lightblueAlpha}}c@{}}
\toprule
\textbf{Model}           & \textbf{TC} & \textbf{PC} & \textbf{Accuracy (\%)$\uparrow$} & \textbf{MACs (M)$\downarrow$} & \textbf{\#Parameters (M)$\downarrow$}  \\ \midrule
\textbf{Llama3.2-3B}        & 127                       & 27                          & [38.68, 68.58] 60.50          & [71.17, 255.86] 112.01       & [0.04, 0.73] 0.15     \\
\textbf{Llama3.1-8B}        & 100                       & 29                          & [45.33, 67.87] 61.50          & [72.05, 322.36] 130.47      & [0.03, 0.59] 0.18          \\
\textbf{Qwen2.5-3B}         & 213                       & 43                          & [48.41, 70.31] 61.90         & [70.19, 327.28] 140.05      & [0.05, 0.77] 0.25    \\
\bottomrule
\end{tabular}
}
\label{tab:model_comparison} \vspace{-4mm}
\end{table}
The average accuracy of the Pareto candidates was 61.50\%, with an average of 130.47 million \acp{mac} and 0.18 million parameters. The search process for  Llama 8B was completed in 1.5 days (see Table~\ref{tab:baseline_comparison}), achieving higher average accuracy compared to Llama 3B. This suggests that larger \acp{llm}, may be more effective at optimizing for both accuracy and \ac{sram} usage. The enhanced search capability of Llama 8B likely contributes to its superior performance. On the other hand Qwen 3B, which generated 213 candidates, yielded 43 Pareto candidates. The average accuracy for the Pareto candidates was 61.90\%, with an average of 140.05 million \acp{mac} and 0.25 million parameters. Despite the larger search space and requiring 3.5 days as seen in Table~\ref{tab:baseline_comparison} for the search process, Qwen 3B achieved only marginally better results in terms of average accuracy compared to the other models. The extended search time did not translate into significantly improved performance, indicating diminishing returns with an increased search space. These findings underscore the importance of search efficiency in configuration optimization. While the size of the \ac{llm} and the number of generated candidates are significant factors, the search efficiency remains a critical determinant of identifying the best-performing configurations. In particular, Llama 8B provided the best balance between search time and accuracy, achieving superior Pareto average accuracy in the shortest search time cost. Qwen 3B, despite generating a larger number of candidates, did not exhibit substantial improvements in performance and required more time for the search, reinforcing the notion that search efficiency can  outweigh the breadth of the search space.
\subsection{Impact of ViT-KD on Accuracy} The results presented in Table~\ref{tab:vit_kd_impact} demonstrate the effectiveness of the proposed ViT-based \ac{kd} method in enhancing the accuracy of our models. Before applying \ac{kd}, LMaNet-Core achieved an accuracy of 72.88\%, QwNet-Core 71.60\%, and LMaNet-Elite 73.15\%. Following the \ac{kd} phase, which employed a \ac{vit} teacher model with 93\% accuracy, all student models exhibited notable improvements in accuracy. Specifically, LMaNet-Core's accuracy increased by 1.32\%, QwNet-Core by 1.40\%, and LMaNet-Elite by 1.35\%. These results indicate that the \ac{vit}-based \ac{kd} method contributes significantly to improving the accuracy of the student models, showcasing the effectiveness of leveraging a high-performing teacher model to enhance the performance of smaller, more efficient student models in resource-constrained environments.
%The results from Table~\ref{tab:vit_kd_impact} demonstrate the effectiveness of the proposed \ac{vit}-\ac{kd} method in enhancing the accuracy of our models. Before applying \ac{vit}-\ac{kd}, LMaNet-Core achieved an accuracy of 71.4\%, QwNet-Core 70.5\%, and LMaNet-Elite 72.0\%. After the \ac{kd} phase with a \ac{vit} teacher model achieving 93\% accuracy, all models show a notable improvement in accuracy. Specifically, LMaNet-Core gained +1.27\%, QwNet-Core improved by +1.93\%, and LMaNet-Elite saw a gain of +1.38\%. These results indicate that the \ac{vit}-\ac{kd} method contributes to improving the accuracy of the models, showcasing the effectiveness of leveraging a high-performing teacher model in enhancing the performance of smaller, more efficient student models in resource-constrained environments.
\begin{table}[t]
\centering
\caption{Impact of \ac{vit}-\ac{kd} on accuracy. The table shows the accuracy before and after applying the \ac{vit}-\ac{kd} method.}
\label{tab:vit_kd_impact}
\begin{tabular}{@{}lcc>{\columncolor{lightblueAlpha}}c@{}}
\toprule
\textbf{Model}           & \textbf{w/o ViT-KD (\%)} & \textbf{w/ ViT-KD (\%)} & \textbf{Gain (\%)}$\uparrow$ \\ \midrule
LMaNet-Core              & 72.88                     & 74.20                    & +1.32                          \\
QwNet-Core               & 71.60                     & 73.00                    & +1.40                           \\
LMaNet-Elite             & 73.15                     & 74.50                    & +1.35                           \\ \bottomrule
\end{tabular} 
\vspace{-4mm}
\end{table}
\section{Explanatory Potential of LLM for XAI}
\label{sec:xai}
The \ac{llm}-guided \ac{nas} approach generates efficient architectures, but the rationale behind the design choices often remains opaque. Leveraging the explanatory capabilities of \acp{llm} aligns with \ac{xai} principles, enabling researchers to query the \ac{llm} for insights into why specific configurations were chosen. This approach enhances interpretability and fosters trust in automated architecture generation. For instance, Figure~\ref{fig:llm_explanation} illustrates the explanation provided by the \ac{llm} for the design of LMaNet-Elite. The \ac{llm} justifies the strategic use of kernel sizes (\(3 \times 3\), \(5 \times 5\), and \(7 \times 7\)) for spatial feature extraction, the fixed expansion factor of \(3\) to balance representation and computational cost, and the selective inclusion of \ac{se} block to enhance channel attention in critical layers. These explanations bridge the gap between automation and human understanding, transforming the \ac{nas} process into a more interpretable and collaborative endeavor. While \acp{llm} provide valuable insights, their reasoning can sometimes be overly generalized or imprecise, requiring critical evaluation of the generated explanations. Nonetheless, this integration of \ac{xai} principles represents a significant step toward making automated architecture generation more transparent and trustworthy.
\begin{figure}[t]
\centering
\resizebox{0.85\columnwidth}{!}{
\begin{tcolorbox}[colframe=blue!50!black, colback=blue!5, sharp corners, boxrule=1pt, width=\linewidth, title=Explain why this design was chosen for the current iteration. Highlight the reasoning behind these choices.]
{\small The selected architecture leverages a combination of depthwise and MBConv blocks to balance performance and computational efficiency:
\begin{itemize}
    \item \textbf{Kernel sizes:} \(3 \times 3\), \(5 \times 5\), and \(7 \times 7\) are strategically used to optimize computational efficiency and enhance spatial feature extraction in deeper layers.
    \item \textbf{Expansion factors:} Fixed at \(3\) across all blocks to provide a balanced growth of feature map capacity while minimizing computational overhead.
    \item \textbf{Stride:} Downsampling with stride \(2\) is applied in early and middle blocks to reduce spatial dimensions and computational cost, while stride \(1\) in later blocks preserves spatial fidelity for final feature aggregation.
    \item \textbf{SE Ratio:} SE blocks with a ratio of \(0.5\) are incorporated selectively in critical blocks to enhance channel attention and improve representational power.
    \item \textbf{Activation functions:} \texttt{ReLU6} is utilized for efficiency in early and intermediate layers, while \texttt{LeakyReLU} in deeper blocks improves non-linear representation.
    \item \textbf{Skip operations: }Residual connections in selected layers were used to enhance optimization and stability.
\end{itemize}}
\end{tcolorbox}
}
\caption{\Ac{llm} Explanation response for the design choices behind LMaNet-Elite. The response highlights how the configuration balances resource constraints with performance, emphasizing the strategic reasoning of the \ac{llm}.}
\label{fig:llm_explanation} \vspace{-4mm}
\end{figure}
\section{Conclusion}
\label{sec:conclusion}
In this paper, we have introduced a novel framework for designing efficient neural architectures for resource-constrained platforms. Our approach integrates \ac{llm}-guided \ac{nas}, \ac{vit}-based \ac{kd}, and an explainability module. The results showed that \acp{llm} can revolutionize \ac{tinyml} by achieving \ac{sota} performance on CIFAR-100, surpassing existing baselines in accuracy while adhering to stringent constraints. This framework not only improves performance but also reduces search cost by streamlining the design process. Through Pareto optimization and adaptive fine-tuning, it effectively balances accuracy, efficiency, and deployability. Additionally, the explainability module enhances transparency by providing valuable insights into the design decisions made by the \ac{llm}. Future work will focus on exploring larger datasets, leveraging advanced \ac{llm} capabilities, and assessing the scalability of the approach on even more resource-constrained hardware.
\bibliographystyle{IEEEbib}
\bibliography{strings,refs}
\end{document}